\documentclass[AMA,STIX1COL]{WileyNJD-v2}
\usepackage{moreverb}
\usepackage{float}
\usepackage{subfig}
\usepackage{amssymb}
\usepackage{amsmath}
\usepackage{graphicx}
\usepackage{color}
\usepackage{bm}
\usepackage{array}
\usepackage{multirow}
\usepackage{booktabs}
\usepackage{url}
\usepackage{algorithm}
\usepackage{algorithmicx}
\usepackage{algpseudocode}
\usepackage{algpascal}
\usepackage{epstopdf}
\usepackage{float}
\usepackage{makecell}

\renewcommand{\P}{{\bf P}}

\renewcommand{\S}{{\bf S}}

\usepackage{bbding}
\usepackage{graphicx}

\newcommand\BibTeX{{\rmfamily B\kern-.05em \textsc{i\kern-.025em b}\kern-.08em
		T\kern-.1667em\lower.7ex\hbox{E}\kern-.125emX}}

\articletype{Article Type}%

\received{<day> <Month>, <year>}
\revised{<day> <Month>, <year>}
\accepted{<day> <Month>, <year>}

\begin{document}
	
	\title{Foreground Object Structure Transfer for Unsupervised Domain Adaptation}
	
	\author[1,2]{Jieren Cheng}%
	\author[1]{Le Liu}%
	\author[1,4]{Xiangyan Tang}%
	\author[5]{Wenxuan Tu}%
	\author[6]{Boyi Liu}%
	\author[3]{Ke Zhou}%
	\author[3]{Qiaobo Da}%
	\author[3]{Yue Yang}%
	
	\address[1]{School of Computer Science and Technology, Hainan University, Haikou, 570228, China.}%
	\address[2]{Hainan blockchain technology engineering research center, Haikou, 570228, China.}%
	\address[3]{School of Cyberspace Security, Hainan University, Haikou, 570228, China.}%
	\address[4]{College of Intelligence and Computing, Tianjin University, Tianjin, 300350, China.}%
	\address[5]{College of Computing, National University of Defense Technology, Changsha, 410073, China.}%
	\address[6]{Guangdong-Hong Kong-Macao Joint Laboratory of Human-Machine Intelligence-Synergy Systems, Univerisity of Macau, Macau, 999078, China.}%
	
	\corres{*Le Liu,School of Computer Science and Technology, Hainan University, Haikou, 570228, China.
		\email{le@hainanu.edu.cn}}
	
	
	\authormark{Cheng \textsc{et al}}
	
	\abstract[Abstract]{Unsupervised domain adaptation aims to train a classification model from the labeled source domain for the unlabeled target domain. Since the data distributions of the two domains are different, the model often performs poorly on the target domain. Existing methods align the feature distributions of the source and target domains and learn domain-invariant features to improve the performance of the model. However, the features are usually aligned as a whole, and the domain adaptation task fails to serve the classification, which will ignore the class information and lead to misalignment.In this paper, we investigate those features that should be used for domain alignment, introduce prior knowledge to extract foreground features to guide the domain adaptation task for classification tasks, and perform alignment in the local structure of objects. We propose a method called Foreground Object Structure Transfer(FOST). The key to FOST is the new clustering based condition, which combines the relative position relationship of foreground objects. Based on this conditions, FOST makes the data distribution of the same class more compact in geometry. In practice, since the label of the target domain is not available, we use the clustering information of the source domain to assign pseudo labels to the target domain samples, and then according to the source domain data prior knowledge guides those positive features to maximum the inter-class distance between different classes and mimimum the intra-class distance. Extensive experimental results on various benchmarks ($i.e.$ ImageCLEF-DA, Office-31, Office-Home, Visda-2017) under different domain adaptation settings prove that our FOST compares favorably against the existing state-of-the-art domain adaptation methods.}

	\keywords{Unsupvised domain adaptation, Contrastive learning, Object structure}
	
	\maketitle

	\section{Introduction}\label{sec1}
	Many computer vision tasks have been made significant progress by deep neural networks(DNNs), especially in the field of supervised learning. One of the important reasons for the great success is that DNNs can learn transferable feature representations in large-scale labeled datasets, such as ImageNet \cite{deng2009imagenet}. Unfortunately, the model trained on large-scale labeled dataset is generally sensitive to domain shifts, that is to say it cannot be generalized well to data that not lies inside the training data distribution\cite{torralba2011unbiased}. How to improve the generalization performance of the model has become one of the hottest research topics. One of the possible solutions is to manually mark some samples on the target domain and then use these samples to fine-tune the model. However, marking a number of samples that can fine-tune the model is expensive and time-consuming, which is not suit for every ad-hoc target domain or task. Recent studies have shown that unsupervised domain adaptation can effectively mitigate the domain shift in data distributions\cite{tzeng2017adversarial,saenko2010adapting,ben2010theory,ben2006analysis}. A majority of successful methods rely on domain-level alignment using minimize the discrepancy between the source and target domain in the deep neural network, where the discrepancy is measured by Maximum Mean Discrepancy(MMD)\cite{long2015learning}, Joint MMD(JMMD)\cite{long2017deep} and cross-covariance\cite{long2018conditional}. In addition, domain-level alignment can also be implemented using adversarial learning to learn domain-invariant features. Although these methods have yielded promising results, most of them did not use the label information of the source domain, which belonged to the class-agnostic method. These methods are likely to ignore the finer class specific structure of the samples, which will cause noisy predictions near classifier boundaries. Some subsequent works\cite{haeusser2017associative,wu2018unsupervised,kang2019contrastive,pan2019transferrable} alleviated this problem by assigning pseudo-labels to the target domain samples to perform class-level alignment. However, these methods These methods generally align the deep features as a whole, which is unreasonable. On the one hand, the deep features contain rich semantic information, but less relative position information. On the other hand, the deep features have both positive features conducive to alignment and negative features unfavorable to alignment, leading to negative alignment.
	
	In order to further align the source domain and target domain at the class-level, we propose to perform alignment on the foreground object structural feature representations. In this way, the relative position information and semantic information of features are considered at the same time, and the positive features are aligned according to categories. Thus, better performance is achieved.
	
	In practice, we obtain the structural information of the target by introducing the location information of the shallow features, and then use the prior knowledge to further extract those features in the structural information that are beneficial to the domain adaptation task. Since labels from the target domain are not available, we use a clustering algorithm to assign pseudo-labels to samples from the target domain. After obtaining the pseudo-labels, we perform comparative learning on the foreground target structures of the source and target domains, and the feature representation of the model can be further adjusted by minimizing the Structural Contrastive Loss(SCL). Ultimately, the model can retain more transferable features of the source and target domains. As the center of the target domain becomes more and more accurate, the number of samples considered will keep increasing. Iterative learning strategies can better assist the optimization of algorithms.
	
	The effectiveness of Foreground Object Structure Transfer(FOST) is reflected by improved adaptation accuracy on public UDA benchmarks: ImageCLEF-DA\footnote{http://imageclef.org/2014/adaptation}, Office-31\footnote{https://faculty.cc.gatech.edu/~judy/domainadapt/}\cite{saenko2010adapting}, Office-Home\footnote{https://www.hemanthdv.org/officeHomeDataset.html}\cite{venkateswara2017deep}, and Visda-2017\footnote{http://ai.bu.edu/visda-2017/}. The experimental results show that our method performs favorably against the state-of-the-art UDA methods.
	
	The main contributions of this paper are summarized as follows:
	\begin{itemize}
		\item We emphasize the importance of foreground object structure in domain adaptation tasks and demonstrate that it is beneficial to improve the performance of domain adaptation.
		\item We propose positive feature alignment, that is, using the prior knowledge of the source domain label to enlarge the weight of positive features in the feature graph, and then align the source domain with the target positive features.
		\item We propose a domain adaptation framework named FOST, which can be trained end-to-end by minimizing SCL.
		\item Our method achieves the competitive performance compared to the state-of-the-art on the ImageCLEF-DA, Office-31, Office-Home and Visda-2017 benchmark.
	\end{itemize}

	The rest of this paper is organized as follows. Section \ref{sec2} reviews the related work based on unsupervid domain adaptation. In Section \ref{sec3}, we present the model design and each component of FOST. Section \ref{sec4} conducts experiments and analyzes the results. Finally, we give conclusions in Section \ref{sec5}.
	
	\section{Related work}\label{sec2}
	In this section, we will briefly introduce the related work of unsupervised domain adaptation, and then highlight the differences between our methods and them. The purpose of unsupervised domain adaptation is to improve the classification accuracy of the model in the target domain by using the labeled source domain data with different distribution from the target domain data. Thanks to the development of deep convolution neural network in recent years, many methods based on deep convolution network have been proposed for unsupervised domain adaptation.
	
	\subsection{Unsupervised domain adaptation based on adversarial learning}
	The method based on adversarial learning is a typical unsupervised domain adaptive method. Inspired by the generation of countermeasure network, Ganin et al. Proposed dnna, which uses training a domain classifier to identify which domain the samples come from. The training stops until the domain classifier cannot distinguish the samples from the source domain and the target domain. Therefore, the difference between the feature distributions corresponding to different domains will be minimized. According to this framework, Tzeng et al. proposed adversarial discriminant domain adaptation (ADDA)\cite{tzeng2017adversarial}, which uses a shared domain classifier to align the outputs of two independent domain feature generators.Through the adversarial adaptation method, SIMNET introduced a classifier based on similarity and adversarial loss to enhance the adaptation performance\cite{pinheiro2018unsupervised}. Zhang et al. proposed multiple domain classifiers are used for cooperative confrontation learning\cite{zhang2015multi}. In addition, the adversarial loss with auxiliary loss can make the model learn domain invariant features more effectively. Meanwhile, the previous domain adaptation method based on confrontation learning only focuses on global transformation. Although it can reduce the distribution difference across domains, it destroys the class semantic features in each sample. As a result, the classifier trained in this way can not get ideal results in the target domain.Recently, GTA solved this problem by modifying the auxiliary classifier GAN (ac-gan) and merging class labels and true or false labels\cite{sankaranarayanan2017unsupervised}. In addition to GTA, symnets\cite{zhang2019domain} solves this problem by making the asymmetric design of source and target task classifiers share their layer neurons with them. These methods apply MMD to single branch GAN structure and obtain excellent results.

	\subsection{Unsupervised domain adaptation based on metric}
	Recently, more and more researchers have proposed reducing unsupervised and adaptive distance metrics. The general assumption of this method is that the domain offset can be measured by some distance metric. Therefore, the domain offset can be approximately reduced to minimize the metric-based loss. Indicators such as the maximum average difference, the F-norm of the covariance, and the maximum density divergence are proposed as measures of the shift between the two domains. Affine Glassmann distance and log Euclidean metric\cite{luo2020unsupervised} are applied to unsupervised domain adaptation. In addition to the above indicators, optimal transmission\cite{le2019differentially} has also attracted attention in recent years. Chen used Wasserstein distance\cite{chen2019aggregated} as a metric loss to improve the similarity between features in different domains, and Shao\cite{shao2020domain} proposed to transfer the source samples to the target domain by estimating the transmission map. Generally speaking, MMD\cite{sejdinovic2013equivalence} is a useful and popular non-parametric metric used to measure the domain difference between the source domain and the target domain. Long also found that it is possible to obtain more transferable features by minimizing the multi-core MMD loss\cite{long2015learning}. JAN\cite{long2017deep} uses the joint maximum average difference criterion to evaluate the loss of the learning transmission network. Ding\cite{ding2018deep} explored deep low-rank coding to extract domain invariant features by adding MMD as a domain alignment loss. However, these methods only measure domain differences and ignore differences between classes. Domain adaptation without class-level domain differences transfers the source data to the target domain, leading to negative migration. Recently, SimNet\cite{pinheiro2018unsupervised} solved this problem by learning domain invariant features and classification prototype representations. In addition, CAN\cite{kang2019contrastive} optimizes the network by considering the difference between the intra-class domain and the inter-class domain. The above methods focus on the deep feature matching of the source and target domains, and we recommend using relative position information and aligning the positive features of the source and target domains.
	
	\section{Proposed Method}
	\label{sec3}
	The current unsupervised domain adaptive methods can be roughly divided into two categories, one of which is to learn the domain invariant features in the source and target domains through domain adversarial learning, and the other is to learn domain-invariant features by aligning the source and target distributions by some metrics. Note that our method belongs to the latter to improve the performance of the metric-based domain adaptive model.
	
	In the problem of unsupervised domain adaptation (UDA), we have a labeled source domain $\mathcal{S}  = \{\mathcal{X}_s,\mathcal{Y}_s\} = \{(x_i^s,y_i^s)\}_{i=1}^{n_s}$ with $n_s$ samples, where $\mathcal{S} \sim \mathcal{P}_s$ along with an unlabeled target domain $\mathcal{T}  = \{\mathcal{X}_t\} = \{x_j^t\}_{j=1}^{n_t}$ with $n_t$ samples, where $\mathcal{T} \sim \mathcal{P}_t$, and $\mathcal{P}_s \neq \mathcal{P}_t$. $x^s$ , $x^t$ represent the input data, and $y^s \in \{0,1, \cdots , C-1\}$ denote the source data label of C classes. The task is to train a deep neural network $\varphi$ using $\mathcal{S}$ and $\mathcal{T}$ that can make predictions $\{\hat{y}^t\}$ on $\mathcal{T}$.
	
	We define $\varphi_l(x)$ as the outputs of layer $l \in \mathcal{L}$ in the deep neural network. In this section, we first review the metric-based unsupervised domain adaptation method, and then introduce our structure contrastive loss with prior knowledge. Finally, we will clarify the architecture and training process of the proposed framework.
	
	\begin{figure*}[t]
		\centering
		\includegraphics[width=1.0\textwidth]{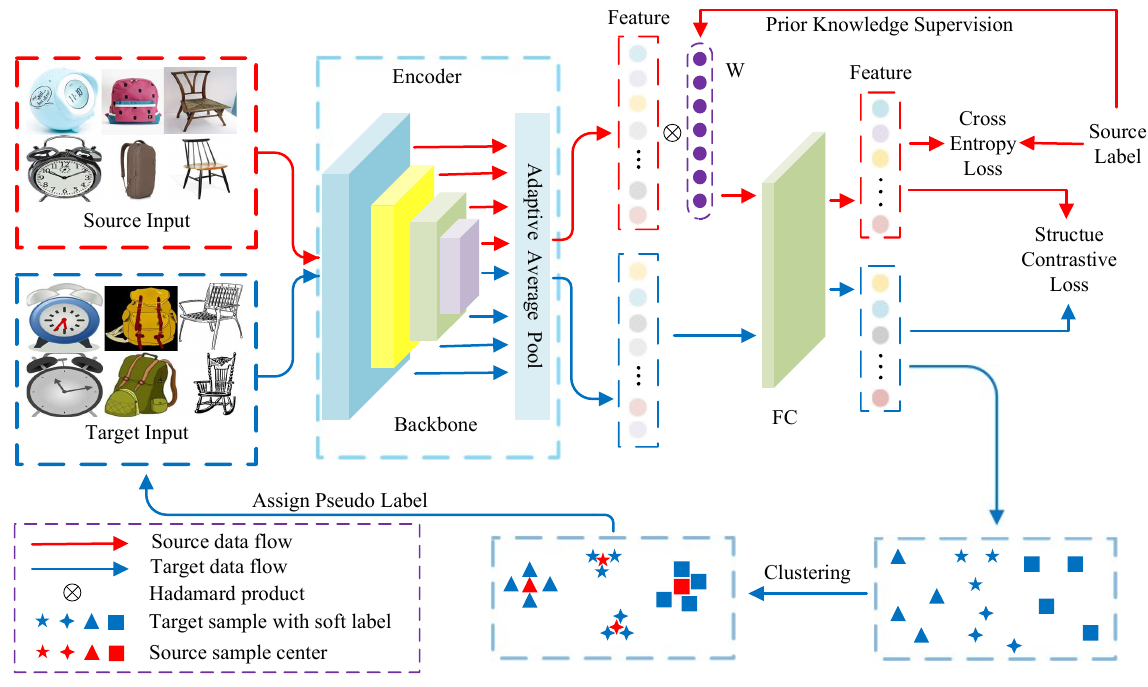} 
		\caption{The framework of Foreground Object Structure Transfer (FOST). Here, an encoder with source and target domain weight sharing is used to extract structural features. Then, the structural features from the source domain are multiplied by the prior knowledge supervision coefficient $\mathcal{W}$  to enhance the proportion of foreground features in the structural features. During the training process, we first minimize the cross-entropy loss through backpropagation to train the encoder. We assign pseudo-labels to the target domain samples using the spherical K-means algorithm. Furthermore we achieve cross domain transfer by minimizing the structural contrastive loss(SCL).}
		\label{figure1}
	\end{figure*}

	\subsection{Overview of Metric-based Domain Alignment}
	
	In order to quantitatively describe the difference between the two data distributions with their mean embedding in the reproducing kernel Hilbert space(RKHS)\cite{mendelson2002learnability}, MMD is usually used as a measurement method. $i.e.$
	
	\begin{equation} 
		\begin{aligned}
			D_{\mathcal{H}}(\mathcal{P}_t,\mathcal{P}_s) 
			=
			\mathop{\sup} \limits_ {f \in \mathcal{H} } (\mathop{E}\limits_{\mathcal{X}_s \sim \mathcal{P}_s}[f(\mathcal{X}_s)]   
			-
			\mathop{E} \limits_ {\mathcal{X}_t \sim \mathcal{P}_t}[f(\mathcal{X}_t)]) 
		\end{aligned}
	\end{equation}
	
	where $\mathcal{H}$ is the RKHS related to the positive definite kernel function. Metric-based domain alignment generally directly minimizeds the MMD(or other metrics) of two domains. However, these methods ignore the relationship between the categories in the two domains, which may cause negative transfer. 
	
	Recently, many studies have proved that if category information is introduced, the performance of domain adaptation can be effectively improved by minimizing the conditional MMD of two domain conditional distributions(need cites). $i.e.$
	
	\begin{equation} 
		\begin{aligned}
			D_{\mathcal{H}}(\mathcal{P}_t,\mathcal{P}_s) 
			=
			\mathop{\sup} \limits_ {f \in \mathcal{H} } (\mathop{E}\limits_{\mathcal{X}_s \sim \mathcal{P}_s}[f(\mathcal{X}_s) | \mathcal{Y}_s]   
			-
			\mathop{E} \limits_ {\mathcal{X}_t \sim \mathcal{P}_t}[f(\mathcal{X}_t) | \mathcal{\hat{Y}}_t ]) 
		\end{aligned}
	\end{equation}
	
	where the label of target domain $\mathcal{\hat{Y}}_t$ is unknown. In this kind of method, the center of the source domain feature distribution is generally obtained as the initialization center of the target domain feature distribution, and then the target domain feature distribution is clustered to assign pseudo labels to the target domain samples.The method in this paper has the same motivation as these methods, but it emphasizes to minimize the discrepancy of foreground object structure distribution between similar samples in the source domain and the target domain, and to maximize the discrepancy of foreground object structure between different classes.
	
	\subsection{Structure Contrastive Loss with Prior Knowledge}
	In previous work, the classification task and the domain adaptation task are often separated. Simply put, the features used for domain adaptation have both positive features related to the classification task and unrelated negative features. Even if a certain metric is reduced to align the source domain distribution with the target domain distribution, there is no obvious benefit to the classification task, and even the existence of negative features will damage the ability to discriminate target features.
	
	The previous work analysis is for classification tasks, the foreground object is highly related to the final classification discrimination information(need cite), and the gradient of the predicted score corresponding to the ground-truth category can effectively convey the classification discrimination information. Inspired by this, we encourage foreground objects structure that are positive features used for classification tasks to obtain greater weight in domain adaptation. In practice, we assign weights to the components of the feature vector to increase the importance of positive features, while the weight of those negative features, usually the background information of a picture, is naturally reduced.

	We propose to take the class information into account, and then perform the operation of minimizing the intra-class foreground objects structure discrepancy and maximizing the inter-class foreground objects structure discrepancy. The proposed \textbf{S}tructure \textbf{C}ontrastive \textbf{L}oss (SCL) is established on the conditional MMD of two domain conditional distributions. We define $f_{stuc}$  as the structure information outputs of the deep neural network, the $pool$ as the adaptive average pool.
	
	the positive feature given by:
	\begin{equation}
		\begin{aligned}
			f_{p} = f_{stuc} \otimes \mathcal{W}
			\label{fp}
		\end{aligned}
	\end{equation}
	where $f_{vec}$ represents the feature map corresponding to the source domain sample $x_s$ through the backbone network and adaptive average pooling $f_{vec} = pool(\tilde {\varphi}(x_s))$.  $\mathcal{W}$ is the gradient of the ground truth label to the feature vector $\mathcal{W} = \partial y / \partial f_{vec}$, $\otimes$ is the hadamard product.supposing $\eta_{cc'}(y,y') = \left\{ \begin{array}{lr}
		1\ if\ y=c,y'=c'; \\
		0\ otherwise.
	\end{array} \right. $ , for two classes $c_1,c_2$ (which can be same or different).

	the $D_{\mathcal{H}}(\mathcal{P}_t,\mathcal{P}_s)$ given by:
	\begin{equation}
		\begin{aligned}
			\hat{D}^{c_1c_2}(\mathcal{\hat{Y}}_t,\tilde{\varphi})
			={\sum _{i=1}^{n_s}}{\sum _{j=1}^{n_s}} \frac{\eta_{c_1c_1}(y_i^s,y_j^s)k(f_{pi}^s,f_{pj}^s)}{{\sum _{i=1}^{n_s}}{\sum _{j=1}^{n_s}}\eta_{c_1c_1}(y_i^s,y_j^s)}+{\sum _{i=1}^{n_t}}{\sum _{j=1}^{n_t}} \frac{\eta_{c_2c_2}(y_i^t,y_j^t)k(f_i^t,f_j^t)}{{\sum _{i=1}^{n_t}}{\sum _{j=1}^{n_t}}\eta_{c_2c_2}(y_i^t,y_j^t)}-2{\sum _{i=1}^{n_s}}{\sum _{j=1}^{n_t}} \frac{\eta_{c_1c_2}(y_i^s,y_j^t)k(f_{pi}^s,f_j^t)}{{\sum _{i=1}^{n_s}}{\sum _{j=1}^{n_t}}\eta_{c_1c_2}(y_i^s,y_j^t)}
		\end{aligned}
	\end{equation}
	
	where $k(f_1,f_2)$ denote the positive definite kernel function. Note that the ground-truth labels of target domain are not available, so we adopt spherical K-means to assign target labels $\mathcal{\hat{Y}}_t$, forming pairs for contrastive learning of multi-level features. Since the clustering algorithm is sensitive to initialization, we set the number of clusters to the number of classes C. For the $c$-th class, the target cluster $O_t^c$ is initialized as the source cluster center $O_s^c$, $i.e. \ O_t^c \leftarrow O_s^c$. Cosine dissimilarity is adopted to assignment of each target sample. After clustering, each target domain samples $x^t$ is associated with a pseudo-label $\hat{y}^t$.
	\begin{equation}
		\begin{aligned}
			dist(f_j^t,O_t^{\hat{y}^t}) < thresholds
		\end{aligned}
	\end{equation}
	where $ dist(x_1,x_2)=\frac{1}{2}(1-\frac{<x_1,x_2>}{\parallel x_1\parallel \parallel x_2 \parallel}) $. $thresholds$ is a constant, samples whose distance from the center is less than $u$ will be assigned pseudo label $\hat{y}^t$. We will remove the ambiguous samples, which are far from its assigned clustered centers to reduce the noise of pseudo-labels. Eq (3) defines two kinds of multi-level features discrepancy, 1) when $c_1=c_2=c$, it measures intra-class foreground objects structure discrepancy; 2) when $c_2 \neq c_2$, it becomes inter-class foreground objects structure discrepancy. The SCL can be given by:
	\begin{equation}
		\begin{aligned}
			\hat{D}^{scl}=\frac{1}{C}{\sum_{c=1}^C}\hat{D}^{cc}(\mathcal{\hat{Y}}_t,\tilde{\varphi})-\frac{1}{C(C-1)}{\sum_{c=1}^C}{\sum_{c'=1 \atop c'\neq c}^C} \hat{D}^{cc'}(\mathcal{\hat{Y}}_t,\tilde{\varphi})
		\end{aligned}
	\end{equation}
	
	The discrepancy in the feature representations of intra-class will be reduced and inter-class will be opposite via minimizing $\hat{D}^{scl}$. Since the training is iterative, which will make the performance of the model continuously improve until convergence.

	\subsection{Foreground Object Structure Transfer}
	
	In deep convolution neural network, the general features extracted by convolution layer are more transferable, while shallow convolution layer will extract high-resolution and weak semantic features, retaining more position information of components in the image, while deep convolution layer is the opposite\cite{long2015learning,long2017deep}. In order to compare the internal structure of the foreground target in the source domain and the target domain, we need to preserve the relative position relationship of each part of the image, that is, the shallow features should be preserved. In practice, we downsample the high-resolution feature map, then merge it with the low-resolution feature map, and finally add it all to the depth feature map to obtain the final feature map. Compared with the feature graph with only high-level semantics, this feature graph retains the internal structural relationship of foreground objects to a greater extent. $i.e.$ the relative position relationship between various organs of the face. This is reasonable. Using pooling and simply concat can preserve the relative position relationship, which is relatively important for domain adaptation. It is worth noting that due to the very high dimension of shallow features, we can not retain all its information, and they can not be used under limited computing resources.
	
	We name our method  Foreground Object Structure Transfer(FOST). The network architecture is shown in Figure \ref{figure1}.
	We need to train the whole network by minimize cross-entropy loss and structure contrastive loss. Since the network has multiple FC layers, the overall cross contrative loss is
	\begin{equation}
		\begin{aligned}
			\mathcal{L}_{scl}=\sum_{l=1}^{L}\hat{D}_l^{scl}
			\label{lscl}
		\end{aligned}
	\end{equation}
	where $L$ represents the number of FC layers. Besides, the network is trained to classify the source images using cross entropy loss $\mathcal{L}_{ce}$
	\begin{equation}
		\begin{aligned}
			\mathcal{L}_{ce}=-\frac{1}{n_s} \sum_{i=1}^{n_s}log P_\theta(y_{i}^{s}|x_{i}^s)
		\end{aligned}
		\label{lce}
	\end{equation}
	where $y^s$ is the ground-truth label of source domain sample $x^s$, $P(y_{i}^{s}|x_{i}^s)$ denotes the predicted probaility of label $y$ with the network parameterized by $\theta$, given input $x$. Combining Eq (5) and Eq (6), the overall objective can be formulated as
	\begin{equation}
		\begin{aligned}
			{\mathop {min} \limits_ {\theta}} {\mathcal{L}} = (1-\alpha)\mathcal{L}_{ce}+\alpha\mathcal{L}_{scl}
			\label{overall}
		\end{aligned}
	\end{equation}
	where $\alpha$ is the weight of the penalty term. As the minimizing ${\mathcal{L}}$, The foreground object structure of the source domain and the target domain will be compared, and finally the intra class domain difference will be minimized and the inter class domain difference will be maximized.
	
	Algorithm 1 shows the training process of FOST. The algorithm is executed in a mini-batch, which may cause the $c$-th samples contained in a mini-batch to come from only one of the source domain or the target domain. Once this happens, the intra-class discrepancy could not be estimated, resulting in reduced training efficiency. The problem solved by class-aware sampling. Simply put,for each mini-batch, only those classes whose number is grater than a constant $N_0$ will be sampled each time target domain samples are assigned.
	
	\alglanguage{pseudocode}
	\begin{algorithm}[!htbp]
		{\caption{The Algorithm of FOST for UDA}\label{Algorithm-proposed}
			\begin{algorithmic}[1]
				\State \textbf{Input}: Source samples  $\mathcal{S}=\{\mathcal{X}_s,\mathcal{Y}_s\}$; Target samples $\mathcal{T}=\{\mathcal{X}_t\}$; Mini-batches $\{B_s,B_t\}$; Training epoch $E$ ; Training iteration $I$.
				\State \textbf{Output}: network parameters $\theta$.
				\State \textbf{while} ${not converge}$ \textbf{do}
				\State \ \ \ \ \ \ $E = 1$.
				
				\State \ \ \ \ \ \ Compute C cluster centers $O_s$ of the S.
				
				\State \ \ \ \ \ \ Initialize the center of $c$-th class  $O_t^c \leftarrow O_s^c$.
				\State \ \ \ \ \ \ Assign pseudo labels to $\mathcal{T}$.
				\State \ \ \ \ \ \ Remove the amibuous samples.
				\State \ \ \ \ \ \ \textbf{for} ${i=1 \rightarrow I}$ \textbf{do}
				\State  \ \ \ \ \ \ \ \ \ \ \ \  class-aware sampling $B_s$ and $B_t$ from $\mathcal{S}$ and $\mathcal{T}$.
				\State  \ \ \ \ \ \ \ \ \ \ \ \  Forward $B_s$ and $B_t$ to get $f_{stuc}$.
				\State  \ \ \ \ \ \ \ \ \ \ \ \  Compute $f_{p}$ using Eq \ref{fp}.
				\State  \ \ \ \ \ \ \ \ \ \ \ \  Compute $\mathcal{L}_{scl}$ using Eq \ref{lscl}.
				\State  \ \ \ \ \ \ \ \ \ \ \ \  Compute $\mathcal{L}_{ce}$ using Eq \ref{lce}.
				\State  \ \ \ \ \ \ \ \ \ \ \ \  Back-propagate with the overall objective Eq \ref{overall}.
				\State  \ \ \ \ \ \ \ \ \ \ \ \  Update network parameter $\theta$.
				\State \ \ \ \ \ \ \textbf{end for}
				
				\State \ \ \ \ \ \ $E = E + 1$.
				\State \textbf{end while}
		\end{algorithmic}}
	\end{algorithm}
	
	\section{Performance Analysis}
	\label{sec4}
	\subsection{Dataset}\ 
	\ \ i) \textbf{ImageCLEF-DA}. The ImageCLEF-DA dataset is a benchmark dataset for ImageCLEF 2014 domain adaptation challenge, which contains three domains: Caltech-256 (C), ImageNet ILSVRC 2012 (I) and Pascal VOC 2012 (P). For each domain, there are 12 categories and 50 images in each category. This dataset consistes of 3,000 images and includes a total of 6 tasks.
	
	ii) \textbf{Office-31}. The Office-31 is a real-wrold dataset, which widely adopted by adaptation methods, containing three distinct domains: Amazon(A), DSLR(D) and Webcam(W). This dataset includes a total of 6 tasks. 
	
	iii) \textbf{Office-Home}. The Office-Home is more challenging benmark dataset, with 65 classes shared by four distinct domains: Artistic images(Ar), Clip Art(Cl), Product images(Pr), and Real-World images(Rw). This dataset includes a total of 12 tasks.
	
	iiii) \textbf{Visda-2017}. The Visda-2017 is a simulation-to-real dataset for domain adaptation with over 280,000 images across 12 categories in the training, validation and testing domains, as with previous work, we evaluated our approach in task: training to validation.

	\subsection{Training Details}
	
	We use pretrained ResNet-50\cite{he2016deep} from ImageNet as the base feature extractor and replace the last fully connected(FC) layer with task-specific FC layer. NVIDIA 3090 GPU used for all experiments in the pytorch framework. Mini-batch stochastic gradient descent(SGD) is used for training the model. The learning rate is the same as the previous work\cite{ganin2015unsupervised,long2015learning,long2017deep}, $i.e.$ the $\eta_p = \frac{\eta_0}{(1+ap)^b}$, the $\eta_0$ is the initial learning rate, p linearly increases from 0 to 1. We set $\eta_0 = 0.001$ for convolutional layers and $\eta_0=0.01$ for FC layer. For all tasks , $a=10$ , $b=0.75$ and $\beta = 0.3$. The $\alpha$ is set to 0.5 for ImageCLEF-DA, 0.25 for Office-31 and Office-Home. The $thresholds$ are set to 1.0 to all tasks of ImageCLEF-DA and Office-Home, 0.05 for Office-31.

	\subsection{Ablation for Structure Feature}
\begin{figure*}[t]
		\centering
		\includegraphics[width=1.0\textwidth]{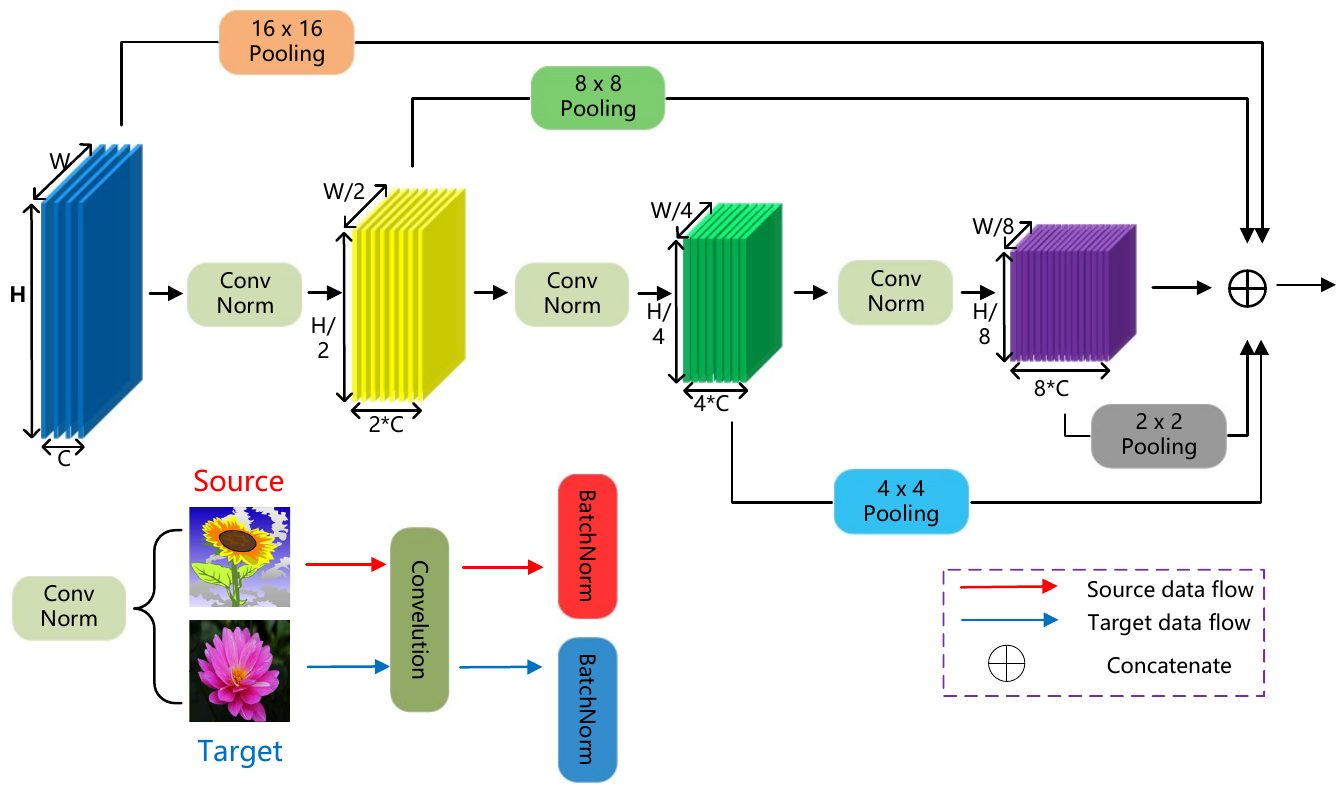}
		\caption{An overview for obtaining the object structure. The shallow feature map with more relative position information will be downsampled through pooled windows of different sizes, and then combined with the deep feature map with more semantic information to generate the object structure features. With the iterative training, the object structure features will be more accurate. Due to the different distribution between the source domain and the target domain, we use the ConvNorm module to normalize the images from the two domains respectively to solve the problem of learning difficulties caused by the change of distribution.}
		\label{figure2}
	\end{figure*}

	In this section, we evaluate the performance of object structure information. As shown in the table \ref{table1}, the mark inside refers to whether or not this level feature is merged into the deepest feature (As shown in the figure \ref{figure2}), we conducted some experiments to study the ablation of object structure information. First, to prove the importance of object structure, we conducted experiments on ImageCLEF-DA. Among them, we can see that compared to the baseline, the effect of each additional level of feature will always be improved. There are two advantages to using object structure information: 1, Since only the downsampling of low-level features is a pooling operation, it will not Increasing the amount of parameters that need to be learned is equivalent to a free upgrade. 2, The low-level features retain the relative position relationship information between the foreground and the background on the larger receptive field. These positions may be useless for classification tasks but useful for domain adaptation tasks, because domain adaptation tasks require alignment operations, The introduction of the location information of low-level features can make the two domains better align.

	\begin{table}[!htbp]
		\begin{center}
			{
				\begin{tabular}{ccccccccccc}
					\toprule
					Model &$S_{2}$&$S_{3}$ & $S_{4}$ &I  $\rightarrow$ P  &P $\rightarrow$ I &I $\rightarrow$ C &C $\rightarrow$ I &C $\rightarrow$ P &P $\rightarrow$ C &Avg  \\
					\midrule
					ResNet-50 & & & &79.0	&93.9	&97.1	&92.6	&78.1	&96.9	&89.6 \\
					$a$ &\checkmark & & &87.0	&98.4	&99.8	&98.3	&86.8	&99.9	&95.0 \\
					$b$ &\checkmark& \checkmark & &87.9	&98.1	&99.1	&98.1	&87.6	&99.9	&95.1 \\
					$c$ &\checkmark &\checkmark& \checkmark  &87.2	&98.0	&99.1	&98.9	&88.1	&99.9	&95.3 \\
					\bottomrule
				\end{tabular}
				\caption{In the ablation study of object structure information, $S_{i}$ represents whether the layer information is introduced. It can be seen that the model with all layers(model c) has better performance than the baseline due to the position information of each part of the image with shallow features.}
				\label{table1}
			}
		\end{center}
	\end{table}
	
	\subsection{Exploration of Generalization Performance.}
	To further demonstrate the generalization of our method, the backbone network is replaced by ResNet-18\cite{he2016deep}. We conduct experiments on six transfer tasks on ImageCLEF-DA, and the result is reported on Table \ref{table2}, showing that FOST has good generalization performance. It is similar to the performance on ResNet-50, both showing that the introduction of target structural information helps to improve performance. Even more impressively, on this dataset, our method even outperforms other methods on the parameter-heavy ResNet-50 on the more lightweight ResNet-18.
	
	\begin{table}[!htbp]
		\begin{center}
			{
				\begin{tabular}{ccccccccccc}
					\toprule
					Model &$S_{2}$&$S_{3}$ & $S_{4}$ &I  $\rightarrow$ P  &P $\rightarrow$ I &I $\rightarrow$ C &C $\rightarrow$ I &C $\rightarrow$ P &P $\rightarrow$ C &Avg  \\
					\midrule
					ResNet-18 & & & &77.3 	&89.2 	&95.8 	&90.5 	&76.8 	&95.7 	&87.6 \\
					$a$ &\checkmark & &	&85.6 	&97.3 	&99.3 	&97.0 	&87.1 	&98.8 	&94.2 \\
					$b$ &\checkmark &\checkmark & &86.5 	&97.0 	&99.5 	&97.3 	&87.5 	&98.7 	&94.4 \\
					$c$ &\checkmark &\checkmark& \checkmark	&87.0 	&98.3 	&99.2 	&97.5 	&87.5 	&99.3 	&94.8 \\
					\bottomrule
				\end{tabular}
				\caption{In the ablation experiment on the generalization performance of FOST, $S_{i}$ indicates whether layer information is introduced. We can see that fost shows robustness to different backbone networks. Similarly, the accuracy of model c is 94.8\%, which is better than the baseline 87.6\%.}
				\label{table2}
			}
		\end{center}
	\end{table}
	
	\begin{figure*}[t]
		\centering
		\includegraphics[width=1.0\textwidth]{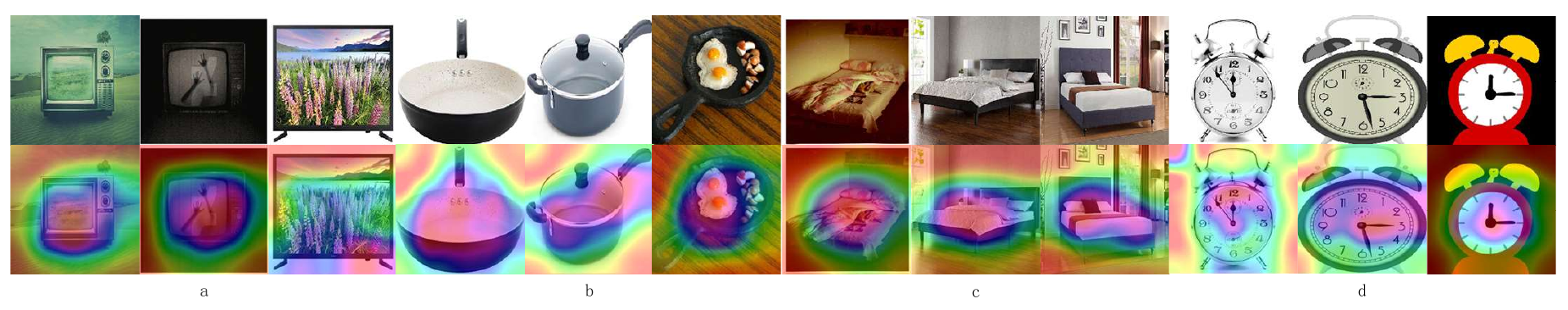}
		\caption{Visualization of foregound object stucture features(best viewed in color). Pictures a, b, c and d represent four categories from the Office-Home dataset: TV, pan, bed and alarm clock. It can be clearly seen that our method focuses on the foreground object of the image ($i.e.$ the purple part in the image) in the comparison between the source domain and the target domain.}
		\label{figure3}
	\end{figure*}
	
	\subsection{Effectiveness of prior knowledge supervision.}
	In order to verify the effectiveness of prior knowledge supervision, we conducted our experiments on the office-home dataset. The last two rows of Table \ref{table5} show the results of the ablation experiment. It can be seen that after the prior knowledge supervision, the accuracy rate has increased by $2.1\%$. Especially in some difficult sub-tasks($i.e$. Cl$\rightarrow$ Ar), there is a more obvious improvement. This shows that the domain alignment task will serve the classification task after the prior knowledge supervision is enabled, so that the model can obtain better performance.In addition, we visualized the feature representation after the use of prior knowledge supervision, as shown in Figure \ref{figure3}, we can see that the network has indeed noticed those foreground features, and these foreground features are generally targets that need to be identified.

	\subsection{Effectiveness of structural contrastive loss}
	We compare the proposed structural contrastive loss with another commonly used metric loss ($i.e.$ MMD loss) to prove the effectiveness of the proposed method. From the Table \ref{table4}, we first observe that the structural contrastive loss is better than the MMD baseline DAN\cite{long2015learning}, which shows that the structural comparison loss based on the idea of positive feature comparison is effective. In addition, we also observed that replacing the structural contrastive loss with the MMD loss caused the accuracy of the relatively simple tasks such as A→D to drop from $93.37\%$ to $90.20\%$, while the accuracy of the difficule tasks such as W→A on the Office-31 dataset dropped from $75.40\%$ to $73.94\%$. From this, we conclude that in order to improve domain adaptability, the performance of using structural contrastive loss is better.

	\subsection{Evaluate in the ImageCLEF-DA Dataset}
	
	The results of ImageCLEF-DA are shown in Table 3. The average accuracy of the model reaches 95.3\%, which is higher than 90.9\% of SRDC\cite{tang2020unsupervised}. It is worth noting that SRDC\cite{tang2020unsupervised}, which assumes structural similarity between the source and target domains, uses a flexible deep network-based discriminative clustering framework to achieve model discrimination. however, SRDC assumes that the source and target domains are identical. It is not always true that the samples of the class correspond to the discriminative clusters that are geometrically close. In our method, we argue that the discriminative clusters of foreground features of the same class in the source and target domains are geometrically more compact, and our method enhances the generality and improves the recognition accuracy. we compared our method on ImageCLEF-DA with typical UDA algorithms: DAN\cite{long2015learning}, DANN\cite{ganin2016domain}, JAN\cite{long2017deep}, CDAN+E\cite{long2018conditional}, TAT\cite{liu2019transferable}, SAFN+ENT\cite{xu2019larger}, SymNets\cite{zhang2019domain}, SRDC\cite{tang2020unsupervised}. We showed the results for adaptation using our method with MLCD loss. We observed that we outperform prior methods by a significant margin. On some relatively simple transfer tasks, such as P$\rightarrow$I, we observed an improvement from 94.7 to 98, indicating the usefulness of our MLCD loss with MlFCNet. Similar improvements can be observed for I$\rightarrow$C, C$\rightarrow$I, P$\rightarrow$C. What's more important is that our method has still made a signigicant improvement in the more difficult transfer tasks I$\rightarrow$P and C$\rightarrow$P, boosts the accuracy of the best method\cite{tang2020unsupervised} by 6.4\% and 8.1\% respectively. On all six transfer tasks of ImageCLEF-DA, the average accuracy is improved by 4.4\% compared with the previous best method.
	
	\begin{table}[!htbp]
		\begin{center}
			{
				\begin{tabular}{cccccccc}
					\toprule
					Method &I $\rightarrow$ P  &P $\rightarrow$ I &I $\rightarrow$ C &C $\rightarrow$ I &C $\rightarrow$ P &P $\rightarrow$ C &Avg  \\
					\midrule
					ResNet-50	&74.8${\pm}$0.3 &83.9${\pm}$0.1 &91.5${\pm}$0.3 &78.0${\pm}$0.2 &65.5${\pm}$03 &91.2${\pm}$0.3 &80.7 \\
					DAN\cite{long2015learning}	&74.5${\pm}$0.4	&82.2${\pm}$0.2	&92.8${\pm}$0.2	&86.3${\pm}$0.4	&69.2${\pm}$0.4	&89.8${\pm}$0.4	&82.5 \\
					DANN\cite{ganin2016domain} 	&75.0${\pm}$0.6	&86.0${\pm}$0.3	&96.2${\pm}$0.4	&87.0${\pm}$0.5	&74.3${\pm}$0.5	&91.5${\pm}$0.6	&85.0 \\
					JAN\cite{long2017deep}	&76.8${\pm}$0.4	&88.0${\pm}$0.2	&94.7${\pm}$0.2	&89.5${\pm}$0.3	&74.2${\pm}$03	&91.7${\pm}$0.3	&85.8 \\
					CDAN+E\cite{long2018conditional}	&77.7${\pm}$0.3	&90.7${\pm}$0.2	&97.7${\pm}$0.3	&91.3${\pm}$03	&74.2${\pm}$0.2	&94.3${\pm}$0.3	&87.7 \\
					TAT\cite{liu2019transferable}	&78.8${\pm}$0.2	&92.0${\pm}$0.2	&97.5${\pm}$0.3	&92.0${\pm}$03	&78.2${\pm}$0.4	&94.7${\pm}$0.4	&88.9 \\
					SAFN+ENT\cite{xu2019larger} &79.3${\pm}$0.1 &93.3${\pm}$0.4	&96.3${\pm}$0.4 &91.7${\pm}$0.0	&77.6${\pm}$0.1	&95.3${\pm}$0.1	&88.9 \\ 
					SymNets\cite{zhang2019domain} &80.2${\pm}$0.3	&93.6${\pm}$0.2	&97.0${\pm}$0.3	&93.4${\pm}$0.3	&78.7${\pm}$0.3	&96.4${\pm}$0.1 &89.9 \\
					SRDC\cite{tang2020unsupervised} &80.8${\pm}$0.3 &94.7${\pm}$0.2	&97.8${\pm}$0.2	&94.1${\pm}$0.2	&80.0${\pm}$03	&97.7${\pm}$0.1 &90.9 \\
					\textbf{FOST(ours)} &\textbf{87.2${\pm}$0.3} &\textbf{98.0${\pm}$0.2} &\textbf{99.1${\pm}$0.1} &\textbf{98.9${\pm}$0.2} &\textbf{88.1${\pm}$0.9} &\textbf{99.9${\pm}$0.1} &\textbf{95.3} \\
					\bottomrule
					
				\end{tabular}
				
				\caption{Accuracy (\%) using Resnet-50 for 6 transfer tasks between three domains of ImageCLEF-DA: Caltech-256 (C), ImageNet ILSVRC 2012 (I) and Pascal VOC 2012 (P). Our method shows state-of-the-art, achieving improvements on all 6 tasks. All methods including ours use ResNet-50 as the backbone architecture.}
				\label{table3}
			}
		\end{center}
	\end{table}

	\subsection{Evaluate in the Office31 Dataset} 
	The results of Office31 are shown in Table 4, for the difficult tasks D→A and W→A where the source domain is small and the domain shift is large. Our model still achieves 78.6\% and 78.8\%, which are higher than 77.5\% and 76.0\% of the best transport model SCAL+SPL. SCAL+SPL uses adversarial learning under an end-to-end structure to maintain intra-class compactness during domain alignment. However, in difficult tasks, because the intra-class compactness is difficult to maintain, compared with SCAL+SPL, we transfer the structure foreground and focus the transfer on the structure of the domain foreground target itself, which means that we can better achieve intra-class performance. of compactness.
	We compared our method on Office-31 with typical domain-level and class-level alignment UDA algorithms: DAN\cite{long2015learning}, DANN\cite{ganin2016domain}, VADA\cite{shu2018dirt}, MSTN\cite{xie2018learning}, MCD\cite{saito2018maximum}, SAFN+ENT\cite{xu2019larger}, iCAN\cite{zhang2018collaborative}, CDAN+E\cite{long2018conditional}, MSTN+DSBN\cite{chang2019domain}, TADA\cite{wang2019transferable}, TAT\cite{liu2019transferable}, SymNets\cite{zhang2019domain}, BSP+CDAN\cite{chen2019transferability}, MDD\cite{zhang2019bridging}, CAN\cite{kang2019contrastive}, SRDC\cite{tang2020unsupervised}, SCAL+SPL\cite{wang2021unsupervised}. We can observe that FOST signigicantly exceeds all compared methods on most taks. The results demonstrate that our FOST achieves better average accuracy than other compared methods, verifying the efficacy of FOST.
	
	\begin{table}[!htbp]
		\begin{center}
			{
				\begin{tabular}{cccccccc}
					\toprule
					Method &A $\rightarrow$W	&D$\rightarrow$W	&W$\rightarrow$D	&A$\rightarrow$D	&D$\rightarrow$A 	&W$\rightarrow$A	&Avg
					\\
					\midrule
					ResNet-50 	&77.8${\pm}$0.2	&96.9${\pm}$0.1	&99.3${\pm}$0.1	&82.1${\pm}$0.2	&64.5${\pm}$0.2	&66.1${\pm}$0.2	&81.1 \\
					DAN\cite{long2015learning} 	&81.3${\pm}$0.3	&97.2${\pm}$0.0	&99.8${\pm}$0.0	&83.1${\pm}$0.2	&66.3${\pm}$0.0	&66.3${\pm}$0.1	&82.3 \\
					DANN\cite{ganin2016domain} 	&81.7${\pm}$0.2	&98.0${\pm}$0.2	&99.8${\pm}$0.0	&83.9${\pm}$0.7	&66.4${\pm}$0.2	&66.0${\pm}$0.3	&82.6 \\
					VADA\cite{shu2018dirt}	&86.5${\pm}$0.5	&98.2${\pm}$0.4	&99.7${\pm}$0.2	&86.7${\pm}$0.4	&70.1${\pm}$0.4	&70.5${\pm}$0.4	&85.4 \\
					MSTN\cite{xie2018learning}	&91.3	&98.9	&\textbf{100}	&90.4	&72.7	&65.6	&86.5 \\ 
					MCD\cite{saito2018maximum} 	&88.6${\pm}$0.2	&98.5${\pm}$0.1	&\textbf{100.0${\pm}$0.0}	&92.2${\pm}$0.2	&69.5${\pm}$0.1	&69.7${\pm}$0.3	&86.5 \\ 
					SAFN+ENT\cite{xu2019larger} 	&90.1${\pm}$0.8	&98.6${\pm}$0.2	&99.8${\pm}$0.0	&90.7${\pm}$0.5	&73.0${\pm}$0.2	&70.2${\pm}$0.3	&87.1 \\
					iCAN\cite{zhang2018collaborative} 	&92.5	&98.8	&\textbf{100}	&90.1	&72.1	&69.9	&87.2 \\
					CDAN+E\cite{long2018conditional} 	&94.1${\pm}$0.1	&98.6${\pm}$0.1	&\textbf{100.0${\pm}$0.0}	&92.9${\pm}$0.2	&71.0${\pm}$0.3	&69.3${\pm}$0.3	&87.7 \\
					MSTN+DSBN\cite{chang2019domain}	&92.7	&99	&\textbf{100}	&92.2	&71.7	&74.4	&88.3 \\
					TADA\cite{wang2019transferable} 	&94.3${\pm}$0.3	&98.7${\pm}$0.1	&99.8${\pm}$0.2	&91.6${\pm}$0.3	&72.9${\pm}$0.2	&73.0${\pm}$0.3	&88.4 \\
					TAT\cite{liu2019transferable}	&92.5${\pm}$0.3	&99.3${\pm}$0.1	&\textbf{100.0${\pm}$0.0}	&93.2${\pm}$0.2	&73.1${\pm}$0.3	&72.1${\pm}$0.3	&88.4 \\
					SymNets\cite{zhang2019domain}	&90.8${\pm}$0.1	&98.8${\pm}$0.3	&\textbf{100.0${\pm}$0.0}	&93.9${\pm}$0.5	&74.6${\pm}$0.6	&72.5${\pm}$0.5	&88.4 \\
					BSP+CDAN\cite{chen2019transferability}	&93.3${\pm}$0.2	&98.2${\pm}$0.2	&\textbf{100.0${\pm}$0.0}	&93.0${\pm}$0.2	&73.6${\pm}$0.3	&72.6${\pm}$0.3	&88.5 \\
					MDD\cite{zhang2019bridging}	&94.5${\pm}$0.3	&98.4${\pm}$0.1	&\textbf{100.0${\pm}$0.0}	&93.5${\pm}$0.2	&74.6${\pm}$0.3	&72.2${\pm}$0.1	&88.9 \\
					CAN\cite{kang2019contrastive}	&94.5${\pm}$0.3	&99.1${\pm}$0.2	&99.8${\pm}$0.2	&95.0${\pm}$0.3	&78.0${\pm}$0.3	&77.0${\pm}$0.3	&90.6 \\
					SRDC\cite{tang2020unsupervised}	&95.7${\pm}$0.2	&\textbf{99.2${\pm}$0.1}	&\textbf{100.0${\pm}$0.0}	&95.8${\pm}$0.2	&76.7${\pm}$0.3	&77.1${\pm}$0.1	&90.8 \\
					SCAL+SPL\cite{wang2021unsupervised}	&\textbf{95.8${\pm}$0.3}	&\textbf{99.2${\pm}$0.4}	&\textbf{100.0${\pm}$0.0}	&94.6${\pm}$0.1	&77.5${\pm}$0.2	&76.0${\pm}$0.2	&90.5 \\
					
					\textbf{FOST(ours)} &95.5${\pm}$0.5 &\textbf{99.2${\pm}$0.1} &\textbf{100.0${\pm}$0.0} &\textbf{96.3${\pm}$0.1} &\textbf{78.6${\pm}$0.3} &\textbf{78.8${\pm}$0.1} &\textbf{91.3} \\

					\bottomrule
					
				\end{tabular}
				\caption{Accuracy (\%) using Resnet-50 for 6 transfer tasks between three domains of Office-31: Amazon(A), Webcam(W), and Dslr(D). Our method shows state-of-the-art, achieving improvements on all 6 tasks. All methods including ours use ResNet-50 as the backbone architecture.}
				\label{table4}
			}
		\end{center}
	\end{table}

	\subsection{Evaluate in the Office-Home Dataset}
	
	The results on Office-Home based on ResNet-50 are reported in Table \ref{table5}, where the results of existing methods are cited from their respective papers. We can observe that FOST outperforms most existing transfer methods by a large margin on all tasks. For the very difficult tasks Ar→Cl, Cl→Ar and Pr→Ar, FOST is improved by 7.1\%, 6.6\% and 9.1\%, respectively. This is reasonable because the four domains contain more categories in Office-Home and the backgrounds of the domains differ greatly, so previous domain adaptation methods are much less accurate. It is encouraging that FOST largely improves current state-of-the-art methods on such a difficult task, which highlights the importance of foreground target structure for domain adaptation.As shown in Table 5,  We compared our method on Office-Home with typical UDA algorithms: DAN\cite{long2015learning}, DANN\cite{ganin2016domain}, JAN\cite{long2017deep}, DWT-MEC\cite{roy2019unsupervised}, CDAN+E\cite{long2018conditional}, TAT\cite{liu2019transferable}, BSP+CDAN\cite{chen2019transferability}, SAFN\cite{xu2019larger}, TADA\cite{wang2019transferable}, SymNets\cite{zhang2019domain}, MDD\cite{zhang2019bridging}, SDRC\cite{tang2020unsupervised}, SCAL+SPL\cite{tang2020unsupervised}. Clearly, we outperformed other approaches by large margins on all tasks.

	\begin{table}[!htbp]
		\begin{center}
			{
				\scalebox{0.8}{
					\begin{tabular}{cccccccccccccc}
						\toprule
						Methods	&Ar$\rightarrow$Cl	&Ar$\rightarrow$Pr	&Ar$\rightarrow$Rw	&Cl$\rightarrow$Ar	&Cl$\rightarrow$Pr	&Cl$\rightarrow$Rw	&Pr$\rightarrow$Ar	&Pr$\rightarrow$Cl	&Pr$\rightarrow$Rw	&Rw$\rightarrow$Ar	&Rw$\rightarrow$Cl	&Rw$\rightarrow$Pr	&Avg \\
						\midrule
						ResNet-50	&34.9 	&50.0 	&58.0 	&37.4 	&41.9	&46.2 	&38.5 	&31.2	&60.4 	&53.9 	&41.2 	&59.9 	&46.1 \\
						DAN\cite{long2015learning}	&43.6 	&57.0 	&67.9 	&45.8 	&56.5 	&60.4 	&44.0 	&43.6 	&67.7 	&63.1 	&51.5 	&74.3 	&56.3  \\
						DANN\cite{ganin2016domain}	&45.6 	&59.3 	&70.1 	&47.0 	&58.5 	&60.9 	&46.1 	&43.7 	&68.5 	&63.2 	&51.8 	&76.8 	&57.6  \\
						JAN\cite{long2017deep} 	&45.9 	&61.2 	&68.9 	&50.4 	&59.7 	&61.0 	&45.8 	&43.4 	&70.3 	&63.9 	&52.4 	&76.8 	&58.3  \\
						DWT-MEC\cite{roy2019unsupervised} 	&50.3 	&72.1 	&77.0 	&59.6 	&69.3 	&70.2 	&58.3 	&48.1 	&77.3 	&69.3 	&53.6 	&82.0 	&65.6  \\
						CDAN+E\cite{long2018conditional} 	&50.7 	&70.6 	&76.0 	&57.6 	&70.0 	&70.0 	&57.4 	&50.9 	&77.3 	&70.9 	&56.7 	&81.6 	&65.8  \\
						TAT\cite{liu2019transferable}	&51.6 	&69.5 	&75.4 	&59.4 	&69.5 	&68.6 	&59.5 	&50.5 	&76.8 	&70.9 	&56.6 	&81.6 	&65.8  \\
						BSP+CDAN\cite{chen2019transferability}	&52.0 	&68.6 	&76.1 	&58.0 	&70.3 	&70.2 	&58.6 	&50.2 	&77.6 	&72.2 	&59.3 	&81.9 	&66.3  \\
						SAFN\cite{xu2019larger} 	&52.0 	&71.7 	&76.3 	&64.2 	&69.9 	&71.9 	&63.7 	&51.4 	&77.1 	&70.9 	&57.1 	&81.5 	&67.3  \\
						TADA\cite{wang2019transferable} 	&53.1 	&72.3 	&77.2 	&59.1 	&71.2 	&72.1 	&59.7 	&53.1 	&78.4	&72.4 	&60.0 	&82.9 	&67.6  \\
						SymNets\cite{zhang2019domain} 	&47.7 	&72.9 	&78.5 	&64.2 	&71.3 	&74.2 	&64.2 	&48.8 	&79.5 	&74.5 	&52.6 	&82.7 	&67.6  \\
						MDD\cite{zhang2019bridging} 	&54.9 	&73.7 	&77.8 	&60.0 	&71.4 	&71.8 	&61.2 	&53.6 	&78.1 	&72.5 	&60.2 	&82.3 	&68.1  \\
						SRDC\cite{tang2020unsupervised}	&52.3 	&76.3 	&81.0 	&69.5 	&76.2 	&78.0 	&68.7 	&53.8 	&81.7 	&76.3	&57.1 	&85.0 	&71.3  \\
						SCAL+SPL\cite{tang2020unsupervised}	&57.3 	&77.5 	&80.7 	&68.8 	&77.9 	&79.3 	&65.2 	&55.9 	&81.7 	&75.0 	&61.0 	&83.9 	&72.0  \\
						FOST(without $\mathcal{W}$) &\underline{62.6} 	&\underline{79.6} 	&\underline{84.8} 	&\underline{71.4} 	&\underline{79.2} 	&\underline{79.8} 	&\underline{69.4} 	&\underline{58.1} 	&\underline{84.5} 	&\underline{75.4} 	&\underline{63.7} 	&\underline{85.7} 	&\underline{74.5}  \\
						\textbf{FOST(ours)} &\textbf{64.4} 	&\textbf{80.9} 	&\textbf{84.2} 	&\textbf{75.4} 	&\textbf{81.9} 	&\textbf{81.6} 	&\textbf{74.3} 	&\textbf{60.4} 	&\textbf{85.1} 	&\textbf{78.8} 	&\textbf{64.9} 	&\textbf{86.9} 	&\textbf{76.6}  \\
						\bottomrule
					\end{tabular}
				}
				\caption{Accuracy (\%) using Resnet-50 for 12 transfer tasks between four domains of Office-Home:Artistic images(Ar), Clip Art(Cl), Product images(Pr), and Real-World images(Rw). Our method shows state-of-the-art, achieving improvements on all 12 tasks. All methods including ours use ResNet-50 as the backbone architecture.}
				\label{table5}
			}
			
		\end{center}
	\end{table}

	\subsection{Evaluate in the Visda-2017 Dataset}

	Table \ref{table6} shows the results on Synthetic to Real transfer task of VisDA-2017. Same as the previous standard UDA setup\cite{long2018conditional}, we use ResNet-101 as the feature extractor for this dataset. The results of Table \ref{table6} shows that FOST is universal for different datasets.
	
	\begin{table}[!htbp]
		\begin{center}
			{
				\scalebox{0.8}{
					\begin{tabular}{cccccccccccccc}
						\toprule
						Method &aireplane	&bicycle	&bus	&car	&horse	&knife	&motorcycle	&person	&plant	&skateboard	&train	&truck	&average \\
						\midrule	
						ResNet-101	&72.30 	&6.10 	&63.40 	&91.70 	&52.70 	&7.90 	&80.10 	&5.60 	&90.10 	&18.50 	&78.10 	&25.90 	&49.37\\ 
						RevGrad\cite{ganin2016domain}	&81.90 	&77.70 	&82.80 	&44.30 	&81.20 	&29.50 	&65.10 	&28.60 	&51.90 	&54.60 	&82.80 	&7.80 	&57.35\\ 
						DAN\cite{long2015learning}	&68.10 	&15.40 	&76.50 	&87.00 	&71.10 	&48.90 	&82.30 	&51.50 	&88.70 	&33.20 	&88.90 	&42.20 	&62.82 \\
						JAN\cite{long2017deep}	&75.70 	&18.70 	&82.30 	&86.40 	&70.20 	&56.90 	&80.50 	&53.80 	&92.50 	&32.20 	&84.50 	&54.50 	&65.68 \\
						MCD\cite{saito2018maximum}	&87.00 	&60.90 	&83.70 	&64.00 	&88.90 	&79.60 	&84.70 	&76.90 	&88.60 	&40.30 	&83.00 	&25.80 	&71.95 \\
						ADR\cite{saito2017adversarial}	&87.80 	&79.50 	&83.70 	&65.30 	&92.30 	&61.80 	&88.90 	&73.20 	&87.80 	&60.00 	&85.50 	&32.30 	&74.84 \\
						SE\cite{french2017self}	&95.90 	&87.40 	&85.20 	&58.60 	&96.20 	&95.70 	&90.60 	&80.00 	&94.80 	&90.80 	&88.40 	&47.90 	&84.29 \\
						\textbf{FOST(ours)}	&\textbf{96.82} 	&\textbf{89.41} 	&\textbf{82.92} 	&\textbf{69.32} 	&\textbf{95.67} 	&\textbf{97.73} 	&\textbf{89.54} 	&\textbf{85.23} 	&\textbf{95.45} 	&\textbf{95.22} 	&\textbf{86.71} 	&\textbf{66.55} 	&\textbf{87.55} \\
						\bottomrule
					\end{tabular}
				}
				\caption{Accuracy (\%) using Resnet-101 for Synthetic to Real transfer task between of VisDA-2017. Our method shows state-of-the-art, achieving improvements on all categories. All methods including ours use ResNet-101 as the backbone architecture.}
				\label{table6}
			}
		\end{center}
	\end{table}

	\section{Conclusion}
	\label{sec5}
	In this work, we improve domain adaptation based on the idea of metric learning. We propose a method based on foreground object structure transfer, FOST. It exploits a multi-sample structural contrast loss to explicitly achieve cross-domain transfer. We show that the structure of foreground objects is very helpful for domain adaptation, while making the geometric distance of different classes larger, while preventing the influence of negative features, usually the background of those images. The proposed method is general and shows competitive results on multiple datasets. Like previous work, our method also has limitations. Due to memory constraints, our proposed method is insufficient to handle the number of samples transferred across domains. Meanwhile, the extraction of foreground features relies on direct guidance from source domain labels, which has limitations. Once the source domain dataset is small, it is difficult to extract enough foreground features for training. In future work, we will investigate how to mitigate our method and handle smaller datasets in the source domain.
	
	\subsection*{Acknowledgements} 
	This work was supported by the Key Research and Development Program of Hainan Province (Grant No.  ZDYF2021GXJS003), the National Natural Science Foundation of China (NSFC) (Grant No.62162024, 62162022 and 61762033), the Major science and technology project of Hainan Province (Grant No.ZDKJ2020012), the Key Research and Development Program of Hainan Province (Grant No.ZDYF2020040), Hainan Provincial Natural Science Foundation of China (Grant Nos.2019RC098) and Hainan Graduate Innovation Project(Qhys2021-189).

	\bibliography{fost}%
	
\end{document}